\documentclass[10pt,lettersize,journal,twoside]{IEEEtran}
\usepackage{algorithm}
\usepackage{algorithmic}
\usepackage{graphicx}
\usepackage{amsmath}
\usepackage{courier}
\usepackage{multirow}
\usepackage[colorinlistoftodos]{todonotes}
\usepackage[colorlinks=true, allcolors=blue]{hyperref}
\usepackage{float}
\usepackage{booktabs}
\usepackage{threeparttable}   
\usepackage{babel}
\usepackage[caption=false]{subfig}
\usepackage{url}
\usepackage{ragged2e}
\usepackage{tabularx}
\usepackage[utf8]{inputenc}
\usepackage{epstopdf}
\usepackage{cite}
\usepackage{xcolor}

\usepackage{blindtext}
\renewcommand{\algorithmicrequire}{ \textbf{Input:}} %Ushttps://www.overleaf.com/project/62493f529716a61230d4f79de Input in the format of Algorithm
\renewcommand{\algorithmicensure}{ \textbf{Output:}} %Use Output in the format of Algorithm
\usepackage{mathptmx}
\DeclareMathAlphabet{\mathcal}{OMS}{cmsy}{m}{n}
\DeclareSymbolFont{largesymbols}{OMX}{cmex}{m}{n}

\begin{document}

\hbadness=1000000000
\vbadness=1000000000
\hfuzz=10pt
\vfuzz=10pt
\title{T3DNet: Compressing Point Cloud Models for Lightweight 3D Recognition}

\author{Zhiyuan~Yang,
        Yunjiao~Zhou,
        Lihua~Xie,~\IEEEmembership{Fellow,~IEEE},
        and Jianfei~Yang,
         % <-this % stops a space
        \thanks{
        	Z. Yang, Y. Zhou, and L. Xie are with the School of Electrical and Electronics Engineering, Nanyang Technological University, Singapore (e-mail: \{yang0674,yunjiao001,elhxie\}@ntu.edu.sg).
        	% This work is supported by NTU Presidential Postdoctoral Fellowship, ``Adaptive Multimodal Learning for Robust Sensing and Recognition in Smart Cities'' project fund, in Nanyang Technological University, Singapore.
        	}
         \thanks{J. Yang is with the School of Mechanical and Aerospace Engineering, Nanyang Technological University, Singapore. J. Yang is the corresponding author (yang0478@ntu.edu.sg).}
}


\markboth{}%
{Yang \MakeLowercase{\textit{et al.}}: T3DNet: Compressing Point Cloud Models for Lightweight 3D Recognition}

\IEEEtitleabstractindextext{%
\begin{abstract}
3D point cloud has been widely used in many mobile application scenarios, including autonomous driving and 3D sensing on mobile devices. However, existing 3D point cloud models tend to be large and cumbersome, making them hard to deploy on edged devices due to their high memory requirements and non-real-time latency. There has been a lack of research on how to compress 3D point cloud models into lightweight models. In this paper, we propose a method called T3DNet (Tiny 3D Network with augmEntation and disTillation) to address this issue. We find that the tiny model after network augmentation is much easier for a teacher to distill. Instead of gradually reducing the parameters through techniques such as pruning or quantization, we pre-define a tiny model and improve its performance through auxiliary supervision from augmented networks and the original model. We evaluate our method on several public datasets, including ModelNet40, ShapeNet, and ScanObjectNN. Our method can achieve high compression rates without significant accuracy sacrifice, achieving state-of-the-art performances on three datasets against existing methods. Amazingly, our T3DNet is 58 times smaller and 54 times faster than the original model yet with only 1.4$\%$ accuracy descent on the ModelNet40 dataset.
\end{abstract}

}

\maketitle

% is not selected <OR> if conference mode is selected - because compsoc
% conference papers position the abstract like regular (non-compsoc)
\IEEEdisplaynontitleabstractindextext

% \fi
\IEEEpeerreviewmaketitle

\ifCLASSOPTIONcompsoc
\IEEEraisesectionheading{\section{Introduction}\label{sec:introduction}}
\else
\section{Introduction}
\label{sec:introduction}
\fi
\IEEEPARstart{R}{ecent} years witness the rapid growth of autonomous driving and 3D sensing~\cite{chen2017multi}. In these fields, 3D point cloud obtained from LiDAR or RGB-D cameras has been brought to increasing attention. Different from 2D images, 3D point clouds capture the geometric information of the objects or the environment~\cite{urmson2008autonomous,huang2017visual}. With the rise of deep learning, numerous deep 3D models proposed around 3D object classification and detection have achieved remarkable success~\cite{elbaz20173d,shi2020pv,zheng2021se}. However, due to the complexity of both point cloud data and its models, it remains a challenge to deploy these high-performance networks on memory- and latency-sensitive edge devices~\cite{lin2020mcunet,yang2020mobileda}. For example, Apple has installed a Lidar to iPhone 12 Pro, enabling mobile 3D sensing and future metaverse applications~\cite{luetzenburg2021evaluation}, but deploying state-of-the-art deep 3D models on iPhone requires low memory and computation complexity.

Some existing studies proposed several solutions to deal with efficient 3D sensing using deep learning models. For example, voxel-based models can use low-resolution voxelization or conduct feature extraction in a constrained structure e.g., octree~\cite{wang2017cnn} to restrain the computational costs. Meanwhile, point-based models use sparse convolution or other operations to speed up the inference, such as FG-Net~\cite{liu2022fg}, submanifold~\cite{graham2017submanifold}, SECOND~\cite{yan2018second}, SPVConv~\cite{tang2020searching}. These works are essentially designing efficient inference components to speed up the models. Although they effectively accelerate the point cloud model by efficient inference components, they can not compress the model size to a significantly small level. A 3D model compression method that reduces computation operations, as well as model size, is highly demanded.

\begin{figure}[tb!]
\centering
\includegraphics[width=1.\linewidth]{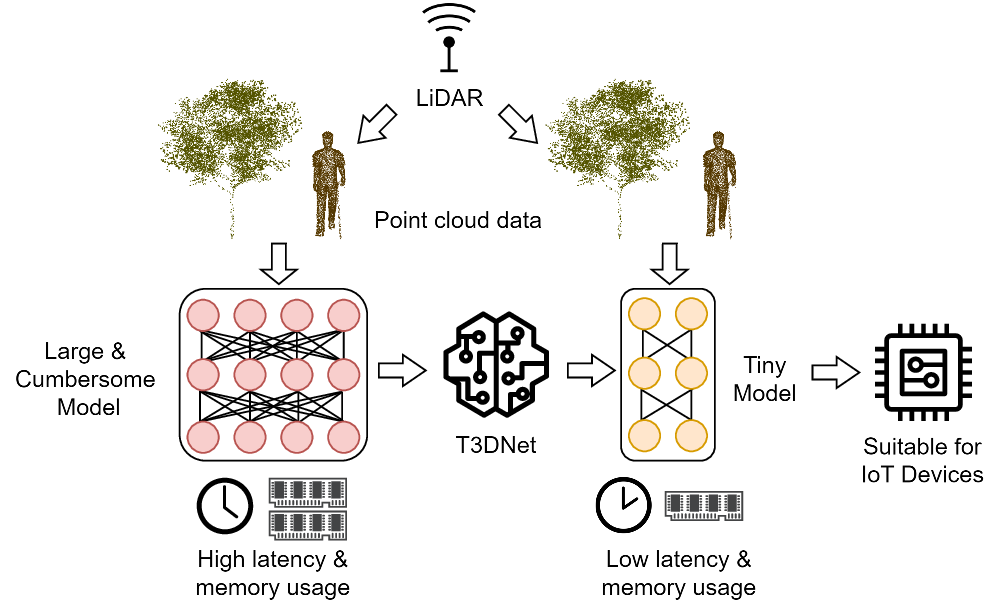}
\caption{T3DNet helps to compress a large 3D point cloud model into a small one with little accuracy drops. The tiny model is more suitable for AIoT (Artificial intelligence of things) devices such as micro-controllers in autonomous cars and other 3D perception sensors. }
\label{fig:concept} 
\end{figure}
\IEEEpubidadjcol
It is challenging to directly compress the deep point cloud network. Most of the prevailing compression techniques in image models, such as pruning~\cite{han2015deep,cui2021joint,hassibi1992second}, and quantization~\cite{gong2014compressing,wu2016quantized,banner2019post}, are unstructured and may significantly affect the model structure. Due to the intrinsic sparsity of the point cloud, an irregular architecture could result in poor generalization of point cloud models. Therefore, we intend to realize point cloud model compression by developing a structured method. Our objective is to obtain a tiny version model of the original one, and try to improve the accuracy of the tiny model. As the tiny model tends to be under-fitting, we expand the tiny model’s capacity and representational ability with a novel network augmentation strategy. Moreover, a common structured compression strategy is knowledge distillation (KD)~\cite{hinton2015distilling}. The tiny model (student) improves its performance by learning from the original model’s guidance (teacher). Although KD has been well leveraged in image models, it is still unclear how to efficiently apply it to point cloud models.

In this paper, we propose T3DNet (Tiny 3D Network with augmEntation and disTillation), a model compression technique for point cloud models. This is inspired by our new observation that the tiny 3D point cloud model after network augmentation is easier for a powerful teacher to distill. Specifically, for a large and cumbersome model, we initialize a tiny version model out of it, maintaining its architecture, and only scaling the network layers. By doing that, the model’s memory and FLOPs can be constrained to an acceptable level. Then we use T3DNet to improve the tiny model’s accuracy. T3DNet is designed to be a two-stage method. In stage 1, we enhance the tiny network’s representational ability by inserting it into a larger model, to add auxiliary supervision by forcing the tiny model to work as a subnet of an augmented network~\cite{cai2021network}. In stage 2, we further improve the performance by knowledge distillation using the original model as a teacher. Extensive experiments are conducted on 3 different datasets, including ModelNet40~\cite{wu20153d}, ScanObjectNN~\cite{uy2019revisiting}, and ShapeNet~\cite{chang2015shapenet}, and the classic point cloud baseline PointNet++~\cite{qi2017pointnet++}. The results show the superiority of our method, and a thorough ablation study is provided to explore the feasibility of an end-to-end strategy and other distillation methods~\cite{romero2014fitnets,zhang2018deep}.

The major contribution of this paper lies in three aspects:
\begin{itemize}
\item We propose a structured compression technique, T3DNet, for the 3D point cloud model. To the best of our knowledge, this is the first work to compress the point cloud model size to a small level without significantly hurting the performance.
\item We discover that the tiny model after network augmentation is easier for a powerful teacher to distill, and design a two-step strategy to further improve the compressed model.
\item Extensive experiments show that our T3DNet achieves state-of-the-art performances on three public datasets against existing methods. On the ModelNet40 dataset, we compress the PointNet++ by 58$\times$ smaller and 54$\times$ faster while only 1.4$\%$ accuracy drops.
\end{itemize}

The rest of this paper is organized as follows: we discuss the related work on point cloud models and knowledge distillation in Section~\ref{sec:related work}. In Section~\ref{sec:method}, we introduce our proposed T3DNet method. After that, we present and analyze the experimental results of our proposed T3DNet with a thorough ablation study on the design in Section~\ref{sec:experiments}. Finally, we conclude the paper and propose our future work in Section~\ref{sec:conclusion}.


\section{Related work}
\label{sec:related work}
\subsection{Point Cloud Models}
There are two main categories of models specialized for point-cloud learning based on how to aggregate point cloud: voxel-based methods~\cite{maturana2015voxnet,riegler2017octnet,wang2017cnn} and point-based methods~\cite{qi2017pointnet,qi2017pointnet++,liu2019relation,komarichev2019cnn}. Voxel-based methods transform the unstructured point cloud into a specific voxel modality, enabling researchers to leverage prior model designs in this domain. These methods typically convert each point cloud into a volumetric occupancy grid representation, i.e., they map each point with coordinates $(x,y,z)$ to a voxel point $(i,j,k)$. Point-based methods, on the other hand, attempt to learn features directly from individual points. Depending on the choice of network architecture, point-based methods can be divided into four main categories and other non-typical methods. 

PointNet~\cite{qi2017pointnet} is a pioneering architecture that uses multi-layer perceptions (MLPs) to learn and classify features. The $n\times3$ 2D input is first processed by a mini-T-Net to generate an affine transformation matrix that addresses rotation-invariance. The model then uses a group of MLPs to extract features and aggregates a global feature using a symmetric aggregation function (such as max-pooling) to address the permutation-invariance problem. Finally, the model classifies the features using a traditional MLP to predict the shape class or point class. However, the drawback of PointNet is that it does not capture local structural information effectively. 

Motivated by this limitation, the hierarchical network PointNet++~\cite{qi2017pointnet++} was proposed to capture local geometric structures from the neighborhood of each point. In this work, the local representation of a point $x$ is defined by the points within a fixed radius sphere centered at $x$. The feature level component set abstraction includes a sampling layer, a grouping layer, and a PointNet layer; several set abstraction layers are stacked together to create a hierarchical network. Another key design feature of PointNet++ is its robust feature learning under non-uniform sampling density. Since point clouds often have a sparse density in different parts, PointNet++ introduces the concepts of multi-scale grouping (MSG) and multi-resolution grouping (MRG) to address this issue. MSG captures patterns at different scales of grouping and concatenates them as the overall feature, while MRG concatenates features from different levels of the set abstractions to generate features of different resolutions. Due to their simplicity and strong performance, PointNet and PointNet++ have inspired many subsequent point-based models.

\textcolor{black}{In recent years, various lightweight point cloud model architectures have garnered attention in the research community. One such framework is 3QNet~\cite{huang20223qnet}, a learning-based point cloud compression approach that dissects point clouds into multiple patches for parallel processing. Each patch undergoes individual compression and encoding through a learned codebook within the Hierarchical Compression framework. Additionally, models like LAM~\cite{cui2020lightweight}, SA-CNN~\cite{puang2022hierarchical}, and DBAN~\cite{zhu2022point} incorporate attention modules to capture both global and local features among unordered 3D points, striking a balance between performance and complexity in point cloud models. While these methods aim to design lightweight point cloud models through compression or convolution modules for efficient feature capture, they fall short of achieving a significantly compact level of model compression.}
\subsection{Model Compression}
\textcolor{black}{In the domain of deep learning, model compression serves the purpose of condensing complex models for deployment on resource-constrained IoT or edge devices like mobile phones, characterized by limited memory and computational resources, while aiming to minimize any significant drop in accuracy. Consequently, a considerable array of model compression and acceleration techniques have been developed, proving highly successful in this pursuit. These methodologies can be classified into several perspectives based on their inherent characteristics, notably encompassing: (1) network pruning and parameter sharing; (2) quantization and binarization; and (3) knowledge distillation.}

\textcolor{black}{Network pruning involves mitigating redundancy within neural networks, adaptable across various architectural levels to achieve differing degrees of compression. Early techniques, exemplified by the OBD pruning algorithm introduced by Jordan and Jacobs~\cite{jordan1998advances}, operated on a neuron-by-neuron basis, while subsequent methodologies like ThiNet~\cite{luo2017thinet} targeted filter-level reductions by removing redundant filters based on approximate outputs compared to the original layer. More aggressive strategies, such as the GAP techniques proposed by Lin et al.~\cite{lin2013network}, replaced dense layers with global mean pooling at the layer level.}

\textcolor{black}{The primary challenge in pruning lies in identifying and exploiting redundancy within neural networks. Initial approaches involved connection reduction based on the Hessian of the loss function~\cite{hanson1988comparing,jordan1998advances,hassibi1992second}. Recent research has unveiled the regularization potential of pruning, sparking interest in models constrained by sparsity. Typically, these constraints leverage $l_0$ or $l_1$ norm regularizers; for instance, straightforward methods prune "0" parameters to meet $l_0$ norms~\cite{lebedev2016fast,zhou2016less}. Structured sparsity learning, as explored by Wen et al.~\cite{wen2016learning}, introduces a regularizer on each level to systematically prune connections, channels, or entire layers. Techniques like $l_1$-norm-based filtering by Li et al.~\cite{li2016pruning} have been employed to efficiently prune ConvNets, with a general preference for $l_1$ or $l_2$-norm regularizers in filter-level pruning endeavors.}

\textcolor{black}{Quantization is a technique involving the mapping of continuous floating numbers onto fixed discrete values, essential for preserving computational accuracy while minimizing the required number of bits. The overarching objective is to optimize bit representation without compromising computational precision. One of the quantization methods is weight clustering, as demonstrated by Gong et al. and Wu et al.~\cite{gong2014compressing,wu2016quantized}, wherein the k-means algorithm clusters parameter values, enabling weights falling within the same cluster to share identical values. Alternatively, approaches that scale the number of bits required to represent each weight have shown promise. Vanhoucke et al.~\cite{vanhoucke2011improving} indicate that quantizing 32-bit parameters to 8-bit equivalents yields significant speed enhancements with minimal accuracy loss. Subsequent studies have further reduced parameters to 16-bit ~\cite{gupta2015deep}, 4-bit~\cite{banner2019post}, or, in extreme cases, employed 1-bit weight representations~\cite{courbariaux2015binaryconnect,rastegari2016xnor}. Merolla et al.~\cite{merolla2016deep} demonstrate the resilience of backpropagation to 1-bit weight distortions, thereby enabling weight binarization within neural networks.}

\textcolor{black}{While network pruning and quantization techniques offer commendable compression performance, most of these methods fall under unstructured compression techniques, potentially resulting in irregular network architectures. Given the inherent sparsity and irregularity of point clouds, the application of these techniques to point cloud models could significantly impact the model's generalization ability. Hence, the development of structured compression techniques tailored for point cloud models becomes imperative.}
\subsection{Knowledge Distillation}
The basic idea of knowledge distillation (KD)~\cite{hinton2015distilling} is to transfer knowledge from a large, powerful teacher network to a small, lightweight student network by learning the soft class distributions after the SoftMax layer. The student network then mimics the teacher model to improve its performance. Deploying deep learning models on embedded systems or mobile robots can be challenging due to their limited computational capacity and memory. Knowledge distillation offers a solution to this problem by allowing us to carefully allocate resources such as memory and computation. For real-time human-robot interaction applications, knowledge distillation is a promising research focus~\cite{zhang2019efficient,kao2021specific,wei2021incremental}.

\begin{figure*}[tb!]
\centering
\includegraphics[width=.9\linewidth]{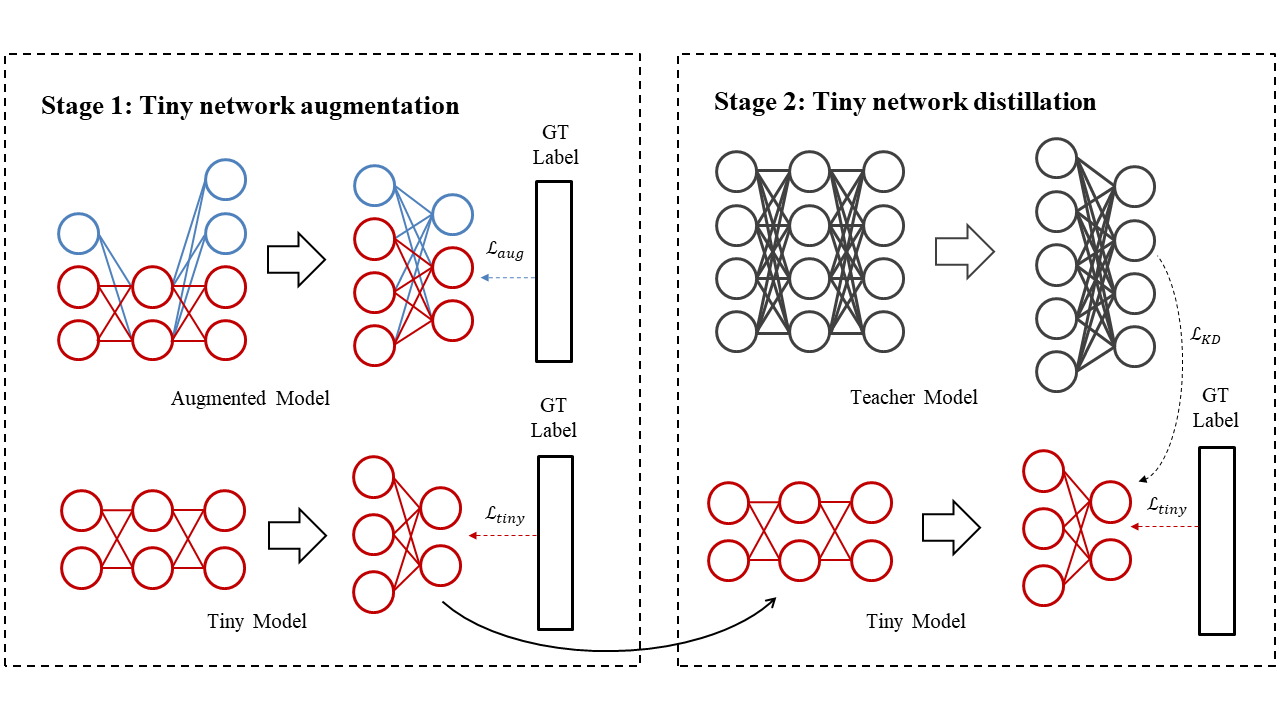}
\caption{Two-stage T3DNet framework. We initialize a tiny model out of the large model. In stage 1, we introduce additional supervision from tiny network augmentation to enhance the tiny model's representative ability. In every epoch, the augmented model is obtained by independently expanding the size of each layer following the expanding ratio strategy. Blue neurons denote the randomly augmented neurons in every epoch. In stage 2, we distill the tiny model by the original network. The augmentation and distillation are conducted in different stages. }
\label{fig:Two-stage-T3DNet} 
\end{figure*}

Logit-based knowledge typically refers to the output of the softmax layer of the teacher model. Early, or "vanilla," knowledge distillation methods used the logits of a deep, complex model as "dark knowledge"~\cite{hinton2015distilling}. Logit-based knowledge distillation is a simple but effective approach for achieving good performance in compression tasks and has gained popularity in solving various tasks and applications. The most popular logit-based knowledge for classification tasks is the soft target~\cite{hinton2015distilling,ba2014deep}. In this approach, the "dark knowledge" is usually evaluated using the Kullback-Leibler divergence loss, which minimizes the difference between the student and teacher logits. The vanilla approach to KD involves joint training of the distillation and student losses. The student loss is typically set as the cross-entropy loss between the ground truth and the softmax class prediction of the student model.

One issue with logit-based distillation is that it can be difficult for very deep networks to learn the logit knowledge, and it is also limited to supervised learning. To address these limitations, the concept of representation learning~\cite{bengio2013representation} can be applied to knowledge distillation. In representation learning, multiple levels of features with increasing abstraction are learned, which can be particularly helpful for thinner and deeper networks in distillation. FitNet~\cite{romero2014fitnets} was the first to introduce the concept of intermediate representation in knowledge distillation, using the hidden layers of the teacher model to supervise the learning of the student model. This type of knowledge, derived from feature activations, is known as "hint". Inspired by this, Zhu \textit{et al.}~\cite{zhu2017feature} developed a feature-distilled network for model compression on the visual tracking system. Additionally, Zhao \textit{et al.}~\cite{zhao2020highlight} proposed using an expert teacher to help the student focus on critical regions through intermediate-level attention maps.

Online distillation refers to a KD approach in which there is no strict definition of teacher and student; rather, two or more peer networks are updated simultaneously, and the entire KD architecture is an end-to-end strategy. In this approach, any network can act as both the student and the teacher for the other model during training, as proposed by Deep Mutual Learning~\cite{zhang2018deep}. Other online distillation methods have been developed, such as the approach of Guo \textit{et al.}\cite{guo2020online}, which extends knowledge by using an ensemble of soft logits, and the approach of Chen \textit{et al.}\cite{chen2020online}, which introduces an auxiliary peer and a group leader in deep learning. Online distillation has the advantage of being able to make use of efficient parallel computing and can save on training costs for a powerful teacher model.

\section{T3DNet}
\label{sec:method}
\subsection{Overview}

To compress a large point cloud model with little performance sacrificed, we initialize a tiny version model out of the original model and make its performance close to the original model. Unlike the original heavy network that needs to avoid over-fitting, a tiny network is more likely to experience under-fitting due to its weak representational ability. To this end, we need to add auxiliary supervision to improve the tiny model's performance. By doing that, the original point cloud model can be compressed to a tiny version, which has close accuracy to the original version but with much fewer parameters and FLOPs.

Inspired by~\cite{cai2021network}, we aim to introduce the network augmentation supervision signal to the tiny model's learning process. The underlying rationale is to expand the tiny model’s capacity by inserting it into large networks, to force it to work as a subnet of a series of augmented networks. Besides that, we also leverage knowledge distillation~\cite{hinton2015distilling}, whose basic idea is to transfer the teacher’s knowledge to guide the student. Through our studies, we discovered that the tiny model after network augmentation is much easier for a teacher to distill. We formulate this discovery as a two-stage T3DNet method. Stage 1: tiny network augmentation, where we train the tiny model with additional supervision from the augmented networks; Stage 2: tiny network distillation, where knowledge distillation is performed on the tiny model after stage 1. We use the original model to act as the teacher in the distillation process.

Fig.~\ref{fig:Two-stage-T3DNet} shows the structure of our T3DNet framework. In stage 1, we obtain the tiny model trained with tiny network augmentation. Next, in stage 2, we distill this tiny model by taking the original model as a teacher to further improve its performance. The distillation utilizes logit-based prediction layer’s method~\cite{hinton2015distilling,ba2014deep}, which outperforms other distillation methods in our studies. We also explore the possibility of an end-to-end strategy. However, rather than introduce these two auxiliary supervision signals in the same stage, we discover the distillation performance is much better after the tiny network augmentation. Knowledge distillation should be introduced with care, blindly merging the distillation in the training process hurts the performance instead of improving it.

\textcolor{black}{In this paper, we leverage PointNet++~\cite{qi2017pointnet++} as the baseline model. PointNet++ is a point-based point cloud model, which extracts the point cloud feature directly from the geometric distribution of the points. In the case of PointNet++, the tiny model is initialized by shrinking the channel sizes in its PointNet~\cite{qi2017pointnet} layer and the classification layers, which utilize Multi-Layer Perceptrons (MLP) as the fundamental learning units. This allows us to effortlessly generate a diminutive model with half or fewer parameters, tailored to specific requirements. In this paper, we find that the tiny model by 1/64 size degradation is promising, so most of the experiments will be conducted on this tiny model.  Furthermore, our T3DNet compression approach is extended to encompass the compression of PointMLP~\cite{ma2022rethinking} and Point Transformer v2~\cite{wu2022point} in Section~\ref{sec:experiments}. This extension is implemented to authenticate the generalization capability of our compression method beyond the confines of PointNet++.}

\subsection{Tiny network augmentation}
Due to the weak representative ability of the tiny model, the accuracy of the tiny model needs to be improved. Therefore, we aim to expand the tiny model’s capacity to ameliorate its representative ability. Inspired by~\cite{cai2021network}, a new learning method aims to tackle the under-fitting problem of the tiny model, we aim to deploy tiny network augmentation for 3D point cloud models. Different from~\cite{cai2021network} whose purpose is achieving accuracy improvement on 2D image models like MobileNetV2-Tiny and MCUNet~\cite{lin2020mcunet}, our method focuses on the 3D point cloud models whose data format is more complex. 

We use weight sharing strategy~\cite{guo2020single,cai2019once} to design and build the series of augmented networks for the tiny model. As is analyzed by~\cite{cai2021network}, it is resource-prohibitive to store different kinds of independent augmented models in the memory. So, we can simplify the structure by storing augmented networks in the largest augmented network. First, we define and store the largest augmented network in memory which contains the whole parameters that could be used. We choose the original PointNet++ model to be the largest augmented network in this paper. The performance of the original model is well-proven, so there is no need for adjusting the size of the largest augmented model. Then the training process is like the reverse version of dropout~\cite{srivastava2014dropout}. We randomly pick an augmented model inside the largest model for each epoch. The method of choosing an augmented network out of the largest model is described below. We define the augmented model by randomly expanding the number of neurons in each tiny network layer to compose different augmented networks. The lower boundary of the augmented layer size is exactly the corresponding tiny model's layer size, and the higher boundary is the size of the largest model in memory. Each layer is augmented independently with a separate expansion ratio. Every randomly picked augmented model will always contain the target tiny model. 

It is complex and unstable to choose the augmented layer size totally randomly, so we constrained the selection to \emph{r} possible expanding ratios. In our studies, we set $\emph{r}=3$, which means there are 3 potential settings for each layer to choose from, ranging from the tiny layer size to the original layer size. For example, assume the tiny layer width is \emph{w}, if the tiny model is a 1/64 degradation model, then the augmented network layer size randomly picks from [\emph{w}, 32\emph{w}, 64\emph{w}] for every training epoch. In this case, the smallest width is \emph{w} which is the size of the tiny model, and 64\emph{w} is the largest width which exactly equals the size of the original model. The medium size 32\emph{w} helps introduce more flexibility to the augmented model. Theoretically, \emph{r} can be any integer value larger than 2 (at least tiny size and original size included), and there is an option to choose different \emph{r} for different layers, however, to keep the strategy simple and effective and to avoid instability, the same $\emph{r}=3$ is set for the whole network.

The total supervision for the learning process comes from two parts, the loss from the tiny model, and the loss from the augmented model. The formulation is below.
\begin{equation}
\mathcal{L}= \beta\mathcal{L}_{tiny}+(1-\beta)\mathcal{L}_{aug}\label{aug}
\end{equation}
, where $\beta$ is a hyper-parameter balancing the supervision between the tiny model and the augmented model. At first, we thought $\beta$ can set to be a static value like 0.5 (total balanced effort) or 0.9 (make training focus on tiny model) for the whole process. But we find it is better to have a different $\beta$ in different stages. We use a linear decay strategy to narrow $\beta$. In the earlier stage, we can use a large $\beta$ like 0.9 to warm up and make the training focus on the tiny model. After the tiny model is trained, we choose a smaller $\beta=0.5$ to introduce more supervision from the augmented models. By doing that, we can help the training model faster to converge.

\subsection{Tiny network distillation}
\label{sec:distillation}
In order to further improve the performance of the tiny network after tiny network augmentation, we aim to apply Knowledge Distillation (KD)~\cite{hinton2015distilling} in the training process. KD involves training a smaller "student" model to replicate the output of a larger "teacher" model. There are various approaches to implementing KD, and it can be challenging to determine the best way to integrate it with tiny network augmentation. Therefore, several studies have been conducted to explore different ways of using both augmentation and distillation to build a tiny network. We will first introduce the KD technique used in our T3DNet and present the results of our ablation studies, which explore different KD techniques and strategies for integrating KD with tiny network augmentation in the next section.

One common way to perform KD is at the prediction layer level~\cite{hinton2015distilling}. In this approach, the student model is trained to mimic the output of the teacher model's softmax layer, known as the "soft label". The difference between the teacher's logit and the student's logit is typically measured using the KL divergence, which is calculated as follows:
\begin{equation}
\mathcal{L}_{KL}(z_t||z_s)=\sum_{i=1}^{N}\sum_{m=1}^{M}z_{t}^{m}(\mathbf{\mathit{x_i} })\log{\frac{z_t^m(\mathbf{\mathit{x_i} })}{z_s^m(\mathbf{\mathit{x_i} })} }  
\end{equation}
where $z_t$ is the soft label from the teacher model and $z_s$ is the logit from the student model, $N$ is the number of samples, $M$ is the data dimension. In many cases, a hyper-parameter called "temperature" ($T$) is introduced to further soften the teacher's knowledge. This leads to the following KD loss function:
\begin{equation}
\mathcal{L} _{KD}=T^{2}\times\mathcal{L} _{KL}(z_t/T||z_s/T) 
\label{kd}
\end{equation}
where $T$ scales the softness of the knowledge. A larger value of $T$ usually leads to a softer label. In our experiments, we tried different values of $T$ from [1, 2, 5, 10, 15] and found that $T=1$ was the best setting for our model. The teacher model in this case is a well-trained PointNet++ model with full parameters. To use this model as the teacher, it is necessary to first train the original PointNet++ network to make it a reliable teacher.

\subsection{Two-stage T3DNet strategy}
\label{sec:combination}
It is difficult to design a method to combine tiny network augmentation with distillation. The simple combination in the training loss cannot always work well and make the model converge. So, it is important to design the training stage for the overall model.  
\begin{algorithm}
	%\textsl{}\setstretch{1}
	\renewcommand{\algorithmicrequire}{\textbf{Input:}}
	\renewcommand{\algorithmicensure}{\textbf{Output:}}
	\caption{Training Procedure of Two-stage T3DNet}
	\label{alg1}
	\begin{algorithmic}[1]
		\REQUIRE teacher model ${W}_{teacher}$
		\STATE \textbf{Stage 1}:
		\STATE Initialize the largest augmented model $W_{L}$ in memory. The tiny model $W_{\text{tiny}}$ can be derived as a subset of $W_{L}$, specifically $W_{\text{tiny}} = W_{L}[0:\text{tiny\_channel}]$.
		\REPEAT
		\STATE Randomly pick the expanding ratio $r$ for each layer to compound the augmented model
		\STATE Optimize the tiny model with the selected augmented model via Equation(1)
		\UNTIL \textit{maximum iteration exhausted}
		\STATE \textbf{Stage 2}:
		\REPEAT 
		\STATE Optimize the tiny model ${W}_{tiny}$ with the teacher model ${W}_{teacher}$ via Equation(3)
		\UNTIL \textit{maximum iteration exhausted}
		\ENSURE The tiny model ${W}_{tiny}$
	\end{algorithmic}  
\end{algorithm}

We design our T3DNet framework as a two-stage strategy, which is simple and intuitive. We train the tiny model with network augmentation in the first stage and distill the tiny model with the original model as a teacher in the second stage. The formulation of this strategy is as below:
\begin{equation}
\begin{split}
Stage \ 1:\mathcal{L}= \beta\mathcal{L}_{tiny}+(1-\beta)\mathcal{L}_{aug}\\
Stage \ 2:\mathcal{L}= \alpha\mathcal{L}_{KD}+(1-\alpha)\mathcal{L}_{tiny}
\end{split}
\end{equation}
Tiny network augmentation and distillation are performed in different stages, and both have tiny model cross-entropy loss as the base supervision. Surprisingly, this strategy is simple but achieves the best performance compared with our other thoughts. Thus, T3DNet is designed to be a two-stage method, with augmentation first followed by knowledge distillation.

In our experiments, we discover the model after the tiny network augmentation is easier for distillation. We train the model with knowledge distillation directly on the pure tiny model from scratch, and the tiny model after network augmentation separately. The results show that for the same training scenario, the distillation of a tiny model with network augmentation is much more effective than simple distillation. This is the core argument that the two-stage T3DNet is a reasonable and effective compression method for the 3D point cloud model.

\section{Experiments}
\label{sec:experiments}
\subsection{Dataset}
\begin{figure*}[htb!]
\centering
\includegraphics[width=5in]{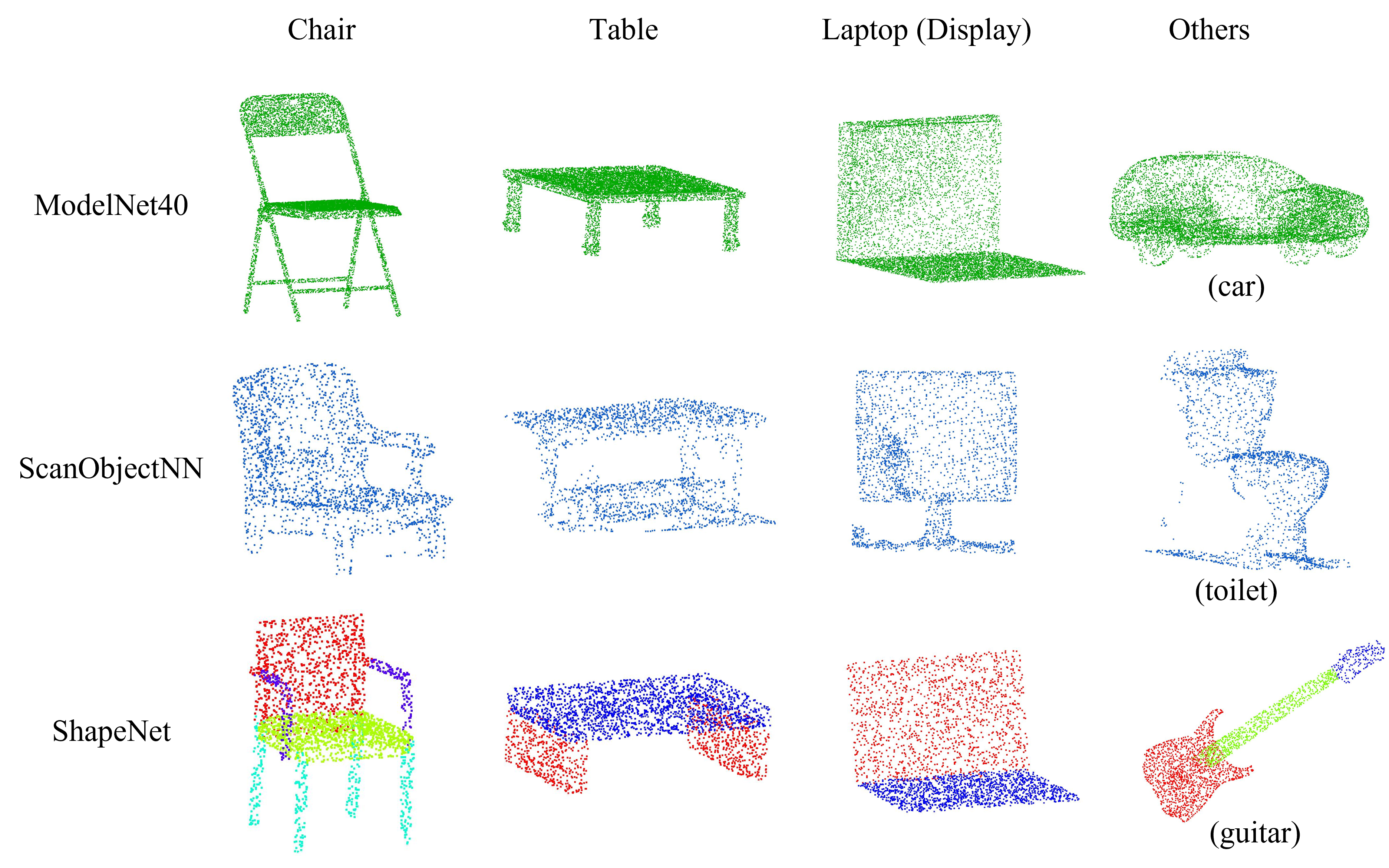}
\caption{Visualization of samples in ModelNet40~\cite{wu20153d}, ScanObjectNN~\cite{uy2019revisiting}, and ShapeNet~\cite{chang2015shapenet}. ModelNet40 is noise-free and clean, while ScanObjectNN is a noisy real-world dataset. ShapeNet is a 3D part segmentation task, whose objects are divided into different parts with the corresponding labels.}
\label{fig:modelnet40} 
\end{figure*}
\textcolor{black}{We conducted our experiments on the four public 3D point cloud datasets. These datasets include two 3D shape classification benchmarks: ModelNet40~\cite{wu20153d} and ScanObjectNN~\cite{uy2019revisiting}; a 3D part segmentation dataset ShapeNet~\cite{chang2015shapenet}, and a 3D semantic segmentation dataset ScanNet v2~\cite{dai2017scannet}. The ModelNet40 dataset consists of 40 categories of clean in-house CAD models and 12,311 CAD samples in total, spilt into 9,843 samples for training and 2,468 samples for testing. Every point has 6 dimensions: 3D geometric positions and 3D face normals. In our experiments, we only use 3D geometric positions as the feature. The ScanObjectNN is a real-world 3D object classification dataset consisting of 2902 real-world instances in 15 classes. Each point cloud has 2048 points. It is more challenging because ScanObjectNN involves noises such as occlusion and background. For the ShapeNet 3D part segmentation task, we use the same split in~\cite{qi2017pointnet}. ShapeNet contains 16,881 objects in 16 classes. Each shape has 2-5 parts, with 50 annotated labels in total. 3D part segmentation is a challenging fine-grained recognition task that involves assigning a part category label to each point. To assess generalization ability, our proposed T3DNet is also tested on PTv2~\cite{wu2022point} for 3D semantic segmentation using ScanNet v2 dataset. The ScanNet v2 dataset contains of 1513 room scans reconstructed from RGB-D data, with 1201 scenes for training and 312 for validation. While we present comprehensive results on ModelNet40, ScanObjectNN, and ShapeNet, our ablation studies and exploratory experiments are exclusively conducted on ModelNet40. The ScanNet v2 dataset is solely employed for validation in the context of compressing PTv2.}

\subsection{Experimental Setups}
\textcolor{black}{In this paper, all experiments are implemented by Pytorch. We use Adam optimizer~\cite{kingma2014adam}, and the scheduler adopts the StepLR strategy. The learning rate was set to 1e-3 and decayed by a factor of 0.7 every 20 steps. The same optimizer and scheduler were used in all experiments. The baseline model for our experiments is the PointNet++ model with multiscale grouping (MSG). We define a tiny model by reducing the channel size of the linear and convolution layers in the original model while keeping other components unchanged. For example, we can derive a tiny model at a 1/64 scale by dividing each channel into 1/8 fractions. This division by 8 on two dimensions of the weight matrix leads to a correspondingly reduced weight matrix size of 1/64. In most experiments, the hyper-parameters were fixed, except for in ablation studies. We set $\alpha=0.5$, $\beta=0.5$, and $T=1$, unless stated otherwise. By default, the models were trained for 200 epochs in Stage 1 and 200 epochs in Stage 2.}

\subsection{Results}
\subsubsection{Effectiveness and Efficiency of T3DNet on point-cloud model}
Table~\ref{table:classification} and Table~\ref{table:partseg} compare the accuracy of the baseline model with our T3DNet method on the ModelNet40, ScanObjectNN, and ShapeNet datasets. There are two baselines, one is the original network, and the other is the pure tiny network without any additional components. The original model has 1.74M parameters and 31.9G FLOPs, while the pre-defined tiny model has only 0.03M parameters and 0.6G FLOPs, representing a compression of 58$\times$ on the number of parameters and 54$\times$ on FLOPs. Despite this significant reduction in size, the T3DNet model only sacrifices 1.45$\%$ in overall accuracy on the ModelNet40 compared to the original model's 92.45$\%$ overall accuracy. On the more challenging ScanObjectNN dataset, the tiny model's overall accuracy is 71.56$\%$, representing a drop of 10.74$\%$ from the original model. However, the T3DNet model is still able to improve the performance of the pure tiny model by 3.38$\%$ to 76.10$\%$.

The experiments on the 3D object part segmentation dataset ShapeNet also show positive results (Table~\ref{table:partseg}). In this task, the original model has 1.74M parameters and 118G FLOPs, while the tiny model has 0.03M \#Params and 2.5G FLOPs. The compression rate is 58$\times$ on \#Params and 46$\times$ on FLOPs. The original model has 94.42$\%$ accuracy and 85.37$\%$ mIoU, while the pure tiny model has 91.25$\%$ overall accuracy and 77.82$\%$ mIoU. T3DNet improves the pure tiny model’s performance by 0.94$\%$ in accuracy and 1.60$\%$ in mIoU. 

Table~\ref{table:classification} and Table~\ref{table:partseg} also demonstrate the effectiveness of our T3DNet design. The experiments are conducted on the pure tiny model, the tiny model with NetAug, and the tiny model with KD. The T3DNet model outperforms all of these models. On the ModelNet40 dataset, the pure tiny model has 88.36$\%$ accuracy, NetAug improves accuracy by 0.38$\%$ to 88.75$\%$, and KD improves accuracy by 1.03$\%$ to 89.39$\%$. However, our T3DNet method can significantly improve the accuracy by 2.64$\%$. The T3DNet model, however, significantly improves accuracy by 2.64$\%$. On the ScanObjectNN dataset, the pure tiny model has 72.72$\%$ accuracy, NetAug decreases accuracy by 1.16$\%$ to 71.56$\%$, KD improves accuracy by 2.62$\%$ to 75.34$\%$, and T3DNet improves accuracy by 3.38$\%$ to 76.10$\%$. These results suggest that the tiny model after network augmentation is better able to learn distillation knowledge from the teacher model. This analysis can also be seen as an ablation study of the T3DNet method, with each component of T3DNet being evaluated individually in the experiments.

\textcolor{black}{In addition to comparing our method with NetAug and KD, we evaluated its performance against two prior knowledge distillation techniques, namely Zhang \textit{et al.}\cite{zhang2020improve} and PointDistiller\cite{zhang2023pointdistiller} in 3D object classification task, as presented in Table~\ref{table:classification}. The results reveal the superiority of our approach over the aforementioned methods. On the ModelNet40 dataset, T3DNet demonstrates a superior overall accuracy, surpassing them by 2.58$\%$ and 2.24$\%$, respectively. Moreover, on the ScanObjectNN dataset, the performance margin widens further, reaching 5.51$\%$ and 5.13$\%$, respectively. This discrepancy in performance could be attributed to the fact that the previous distillation methods excel in scenarios involving small compression rate student models but struggle to generalize effectively when confronted with the significantly higher compression rate in tiny models. Conversely, our proposed method showcases its robust performance even in the face of remarkably high compression rates.}

\textcolor{black}{We evaluated the real-time GPU efficiency of our methods using one RTX3090 GPU on ModelNet40 benchmark, considering both training and testing times per sample as detailed in Table~\ref{table:classification} and Table~\ref{table:partseg}. In the 3D object classification task, our two-stage method T3DNet, exhibits training times of 129.62 ms/sample for Stage 1 and 64.07 ms/sample for Stage 2. For the 3D part segmentation task, the corresponding training times are 49.51 ms/sample for Stage 1 and 26.38 ms/sample for Stage 2. While T3DNet, as a compression technique, inherently demands more training time than the original model, its competitive speed surpasses that of previous KD methods~\cite{zhang2020improve, zhang2023pointdistiller} while maintaining higher accuracy.}

\textcolor{black}{The testing times for the tiny model are 31.45 and 8.96 ms/sample for the classification and part segmentation tasks. Although the testing time does not exhibit a significant reduction as indicated by FLOPs, the tiny model possesses an advantage in terms of the lower parameters it stores in memory.}

\begin{table*}[htbp!]
\centering
\caption{\textcolor{black}{Performance Comparison of T3DNet and Other Methods in 3D Object Classification Tasks.}}
\begin{threeparttable} 
        \begin{tabular}{l|ccc|ccc|c|c|c|c}
            \toprule
                \multirow{2}{*}{Methods} &
                \multicolumn{3}{c|}{\textbf{ModelNet40}} &
                \multicolumn{3}{c|}{\textbf{ScanObjectNN}} &
                \multirow{2}{*}{\#Params} &
                \multirow{2}{*}{FLOPs} &  
                \multirow{2}{*}{Training time} &
                \multirow{2}{*}{Testing time} \\
            \cline{2-7}
             &
            \textbf{OA} & $\Delta \text{Acc}_{1}$ & $\Delta \text{Acc}_{2}$ &
            \textbf{OA} & $\Delta \text{Acc}_{1}$ & $\Delta \text{Acc}_{2}$ & & &
            \\
            
            %{\multirow{2}*{Method}} &  \multicolumn{3}*{ModelNet40} & \multicolumn{3}*{ScanObjectNN} & \#Params & \#FLOPs \\
            %{} & Acc & \Delta Acc_{1} & \Delta Acc_{2} & Acc & \Delta Acc_{1} & \Delta Acc_{2} & {} & {} \\
            \midrule
            Original model (Baseline 1) & 92.45 & -&- & 82.30 & -& -& 1.74M & 31.9G & 65.03 & 32.01\\
            \midrule
            Tiny model (Baseline 2) & 88.36  & -4.09 &-& 72.72 & -9.58 & -& \multirow{6}*{0.03M} & \multirow{6}*{0.6G} & 62.89 & \multirow{6}*{31.45}\\ 
            Tiny model + NetAug~\cite{cai2021network} & 88.75 &  -3.70&+0.38 & 71.56 & -10.74 & -1.16 & & &129.62&\\ 
            Tiny model + KD~\cite{hinton2015distilling} & 89.39 & -3.06 & +1.03 & 75.34 & -6.96 & +2.62 & &  &63.33 &\\ 
            Tiny model + Zhang \textit{et al.}~\cite{zhang2020improve} & 88.43 & -4.02 & +0.07 & 70.59 & -11.71 & -2.13  & &  & 368.71 &\\               
            Tiny model + PointDistiller~\cite{zhang2023pointdistiller} & 88.77 & -3.68 &+0.41  &70.97  & -11.33 & -1.75 & &  &208.16 &\\ 
         
            Tiny model + Two-stage T3DNet & \textbf{91.01} &  -1.45 & +2.64 & \textbf{76.10} & -6.20 & +3.38 & & &129.62/64.07 &\\
            \midrule
            \end{tabular}
            \begin{tablenotes}   
            \footnotesize  
             \item [1] We evaluate our methods by the metrics of \textbf{OA} (overall accuracy, $\%$), \#Params (parameters), FLOPs, Training time ($ms/sample$) and Testing time ($ms/sample$). 
             \item [2] The $\Delta Acc_{1}$ is the accuracy compared with the original model (baseline 1), while $\Delta Acc_{2}$ is the accuracy compared with the tiny model (baseline 2). 
             \item [3] Regarding the training time for Two-stage T3DNet, we present the latency for Stage 1 and Stage 2 separately as $time_1/time_2$ per sample.
            \end{tablenotes}
            % \bottomrule
            \hrule
    % \end{minipage}}
\end{threeparttable} 
\label{table:classification}
\end{table*}

\begin{table*}[htbp!]
\centering
\caption{\textcolor{black}{Performance Comparison of T3DNet and Other Methods in 3D Object Part Segmentation Tasks.}}
\begin{threeparttable} 
        \begin{tabular}{l|cccccc|c|c|c|c}
            \toprule
                \multirow{2}{*}{Methods} &
                \multicolumn{6}{c|}{\textbf{ShapeNet}} &
                \multirow{2}{*}{\#Params} &
                \multirow{2}{*}{FLOPs} &
                \multirow{2}{*}{Training time} &
                \multirow{2}{*}{Testing time}\\
            \cline{2-7}
             &
            \textbf{OA} & $\Delta \text{Acc}_{1}$ & $\Delta \text{Acc}_{2}$ &\textbf{mIoU} & $\Delta \text{mIoU}_{1}$ & $\Delta \text{mIoU}_{2}$ & & &
            \\
            %{\multirow{2}*{Method}} &  \multicolumn{3}*{ModelNet40} & \multicolumn{3}*{ScanObjectNN} & \#Params & \#FLOPs \\
            %{} & Acc & \Delta Acc_{1} & \Delta Acc_{2} & Acc & \Delta Acc_{1} & \Delta Acc_{2} & {} & {} \\
            \midrule
            Original model (Baseline 1) & 94.42 & -&- & 85.37 & -& -& 1.74M & 118G  & 31.19 & 11.35\\
            \midrule
            Tiny model (Baseline 2) & 91.25  & -3.17 &-& 77.82 & -7.55 & - & \multirow{4}*{0.03M} & \multirow{4}*{2.5G} &26.58&  \multirow{4}*{8.96}\\ 
            Tiny model + NetAug~\cite{cai2021network} & 90.81 & -3.61&-0.44 & 77.01  & -8.36 & -0.81 & & & 49.51 &\\ 
            Tiny model + KD~\cite{hinton2015distilling} & 91.72 & -2.70 & +0.47 & 78.83 & -6.54 & +1.01 & & &38.30 &\\ 
            Tiny model + T3DNet & \textbf{92.19} &  -2.23 & +0.94 & \textbf{79.42} & -5.95 & +1.60 & & &49.51/26.38 &\\
            \midrule
            \end{tabular}
            % \bottomrule
    % \end{minipage}}
\end{threeparttable} 
\label{table:partseg}
\end{table*}

\subsubsection{Generalization Across Multiple Architectures}
\begin{table}[htbp!]
\centering
\caption{Validation of T3DNet generalization abilities by compressing PointMLP on ModelNet40 and Point Transformer v2 on ScanNet v2.}
\begin{threeparttable} 
            \begin{tabular}{l|c|c}
            \toprule
            Model & OA &\#Params\\
            \midrule
            PointMLP~\cite{ma2022rethinking}& 94.5 & 12.6M  \\
            \midrule
            Tiny PointMLP & 90.24 &\multirow{4}*{0.21M} \\   
            Tiny model + NetAug& 89.48 &\\
            Tiny model + KD&90.49 &\\
            Tiny model + T3DNet& \textbf{91.75} &\\
            \midrule
            \midrule
            Model & mIoU &\#Params\\
            \midrule
            PTv2~\cite{wu2022point}&75.20 &12.8M  \\ 
            \midrule
            Tiny PTv2&67.64 &\multirow{4}*{0.25M}\\
            Tiny model + NetAug&68.90 &\\
            Tiny model + KD&72.13 &\\
            Tiny model + T3DNet&\textbf{73.46} &\\
            \bottomrule
            \end{tabular}

    % \end{minipage}}
\end{threeparttable} 
\label{table:other_models}
\end{table}
\textcolor{black}{To assess the generalization capabilities of T3DNet across diverse model architectures, we conducted experiments on PointMLP~\cite{ma2022rethinking} and Point Transformer v2 (PTv2)\cite{wu2022point}. PointMLP was validated on the ModelNet40 dataset for the 3D classification task, using overall accuracy (OA) as the evaluation metric. Due to the official implementation of PTv2 utilizing ScanNet v2\cite{dai2017scannet} as the benchmark dataset, we further verified our method on the ScanNet v2 semantic segmentation task, employing mean intersection-over-union (mIoU) as the evaluation metric.}

\textcolor{black}{The corresponding tiny models were generated by reducing the channel size of each layer by 1/8 fractions, similar to the approach used for PointNet++. As depicted in Table~\ref{table:other_models}, our method demonstrated a substantial compression rate while preserving high accuracy and mIoU. Notably, we achieved a 98$\%$ parameter reduction on PTv2 with only a modest sacrifice of 1.74 mIoU. These results underscore the good generalization ability of T3DNet when applied to various point cloud models.}

\subsubsection{Experiments on different tiny model size}
\begin{table*}[htbp!]
\centering
\caption{Tiny network augmentation on different model size}
\begin{threeparttable} 
            \begin{tabular}{l|c|c|c|c|c}
            \toprule
            Model & \#Params  & FLOPs & Test Acc & $\Delta \text{Acc}_{1}$ & $\Delta \text{Acc}_{2}$\\
            \midrule
            Original model& 1.74M & 31.9G & 92.45 & - & -  \\
            \midrule
            1/16 Tiny model & \multirow{2}*{0.11M} & \multirow{2}*{2.2G} &91.76&-&-\\
            1/16 Tiny model + T3DNet & & &91.92 &-0.54 & +0.15 \\ 
            \midrule
            1/64 Tiny model &\multirow{2}*{0.03M}& \multirow{2}*{0.6G}& 88.36&-&-   \\ 
            1/64 Tiny model + T3DNet &  & &91.01  &-1.45 & +2.64 \\ 
            \bottomrule
            \end{tabular}

    % \end{minipage}}
\end{threeparttable} 
\label{table:model_size}
\end{table*}
The trade-off of the compression rate and accuracy loss is also a research focus. Table ~\ref{table:model_size} summarizes the compression and speedup rate for the different tiny models defined. In some cases, it may be necessary to consider a smaller compression rate in order to maintain a stable level of performance. In Table ~\ref{table:model_size}, we use a 1/16 tiny model as a conservative example. The 1/16 tiny model has 0.11M parameters and 2.2G FLOPs, representing a compression of 15.8$\times$ and 14.8$\times$ on the number of parameters and FLOPs, respectively, compared to the original model. Despite this relatively high compression rate, the accuracy loss after applying T3DNet is only 0.54$\%$. This demonstrates that the T3DNet method can maintain a high level of performance even with relatively low compression rates. In this paper, the focus is on exploring the highest possible compression rate without significant performance degradation, and most experiments were therefore conducted using the 1/64 tiny model. However, the T3DNet method can be applied to models with a range of different compression rates.

\begin{table}[htbp!]
\centering
\caption{Comparison of different distillation methods attempts on ModelNet40.}
        \begin{tabular}{c|c|c}
            \toprule
            Model(\#Params=0.03M, FLOPs=589M) & Test Acc &$\Delta$ Acc\\
            \midrule
            Tiny model (Baseline)&88.37 &-   \\
            \midrule
            Tiny model + KD\cite{hinton2015distilling} & 89.39 &+1.09\\
            Tiny model + Hint\cite{romero2014fitnets}& 86.99& -1.37  \\ 
            Tiny model + Mutual Learning\cite{zhang2018deep}&81.25&-7.12   \\ 
            Tiny model + End-to-end T3DNet &  88.35  &-0.02 \\ 
            Tiny model + Two-stage T3DNet &  91.01  &+2.64 \\ 
            \bottomrule
        \end{tabular}
    % \end{minipage}}
\label{table:Distill}
\end{table}

\subsubsection{Exploratory Experiments and Ablation Studies}
\label{sec:exploratory}
We have explored different kinds of distillation and combination methods. 

\textbf{Feature level distillation}: To improve knowledge distillation (KD) at the feature level, the output of a teacher's hidden layer (hint~\cite{romero2014fitnets}) is used. This method is different from prediction layer distillation, as it uses Mean Squared Error (MSE) loss to compute the difference between the teacher and student. To match the size of the student's output with the teacher's output size, a linear mapping layer is added. The formulation for this method is as follows:
\begin{equation}
\mathcal{L} _{hint}=\frac{1}{2} ||u_t(x)-r(u_s(x))||^2
\end{equation}
where $u_t$ is the teacher's feature output, $u_s$ is the student's feature output, $r(\cdot)$ is the linear mapping operation. The parameters of $r(\cdot)$ are also updated during training. However, the results of our experiments (shown in Table~\ref{table:Distill}) indicate that feature-level distillation does not improve the convergence of the model. This is because the hint-based learning method is more suitable for scenarios where the student network is deep and thin, and requires more supervision from the middle layers. The PointNet++ model used in our work does not have this property. This distillation method still requires the original model to work as the teacher.

\textbf{Mutual learning}: Deep mutual learning~\cite{zhang2018deep} is an online distillation method. Different from traditional one-way distillation from a static teacher to a student, mutual learning allows two students to learn from each other collaboratively and train the peer networks in one learning process. The idea of two networks' mutual learning can be applied to the training process of tiny network augmentation, i.e., training the tiny network and augmented network at the same time. Although the purpose of tiny network augmentation is only to obtain a well-trained tiny model, the training process will also make augmented networks converge at last because the model optimizes the whole parameters of the largest model. So we can introduce distillation loss for both the tiny model and augmented model to let them be each other’s teachers in the training stage. The formulation of this process can be shown as:
\begin{equation}
\begin{split}
\mathcal{L} _{KD_1}=T^{2}\times\mathcal{L} _{KL}(z_{org}/T||z_{tiny}/T)\\
\mathcal{L} _{KD_2}=T^{2}\times\mathcal{L} _{KL}(z_{tiny}/T||z_{org}/T)
\end{split}
\end{equation}
where ${L} _{KD_1}$ is the loss for the tiny model, while ${L} _{KD_2}$ is the loss for the augmented model. In this case, we do not need a pre-trained powerful teacher, the augmented model is trained along with the tiny model. We still use KL divergence as the distillation loss, and the training process is end-to-end. However, after the experiments, we find that this method cannot improve the performance of the tiny model.

\textbf{End-to-end strategy}: This strategy involves embedding knowledge distillation in the training process of the tiny network augmentation. The augmentation loss and distillation loss are reduced simultaneously. In the beginning, there is a warm-up stage without distillation loss to make the network converge faster. After a certain epoch ($P_w=100$ in our work), the distillation loss is added until the end of training.

The formulation of this strategy requires precise design. We need scaling factors to control the effort of ground truth supervision, auxiliary augmented supervision and distillation supervision. Thus, two hyper-parameters $\alpha$ and $\beta$ are introduced to the formulation:
\begin{equation}
\mathcal{L}_{end}= \alpha\mathcal{L}_{KD}+(1-\alpha)(\beta\mathcal{L}_{tiny}+(1-\beta)\mathcal{L}_{aug})\label{endtoend}
\end{equation}
where $\mathcal{L}_{KD}$ is the distillation loss, $\mathcal{L}_{aug}$ is the auxiliary loss from the augmented model, and $\mathcal{L}_{gt}$ is the ground truth loss. $\beta$ is the scaling factor to balance between tiny model loss and augmented model loss stated in the previous section. $\alpha$ is controlling the trade-off between the distillation loss and cross-entropy loss. One of the challenges is the setting of these two parameters. The strategy of deciding $\beta$ is described in the previous section. As for $\alpha$, we can set a static value like [0.5, 0.9] or decide the value of $\alpha$ iteratively in the training process. A priori assumption is starting with a large $\beta$ and large $\alpha$ to make the tiny model learn more from the ground truth. Then gradually anneal the scaling factors to amplify the importance of auxiliary augmented and distillation supervision. In our work, we do experiments on the different settings of $\alpha$ and $\beta$ to find the best scaling strategy for the learning process. We also try different distillation methods as described in the previous section for end-to-end learning. However, compared with a two-stage strategy, we find all the end-to-end designs cannot improve the tiny model accuracy significantly.

\textcolor{black}{We attribute the challenges encountered in the end-to-end strategy to the concurrent application of Knowledge Distillation (KD) alongside augmented models. As the augmented models undergo optimization during training, the distilled knowledge permeates across both the augmented models and the target tiny model, introducing a challenge in maintaining focused optimization specifically for the tiny model. Conversely, our T3DNet strategically executes KD after network augmentation. By doing so, the teacher's knowledge is directed solely towards optimizing the target tiny model without simultaneous influence on the augmented models. This deliberate sequencing ensures more precise and dedicated optimization for the tiny model, contributing to the enhanced robustness of the Knowledge Distillation process in the second stage.}
\begin{figure}[tb!]
\centering
\includegraphics[width=0.8\linewidth]{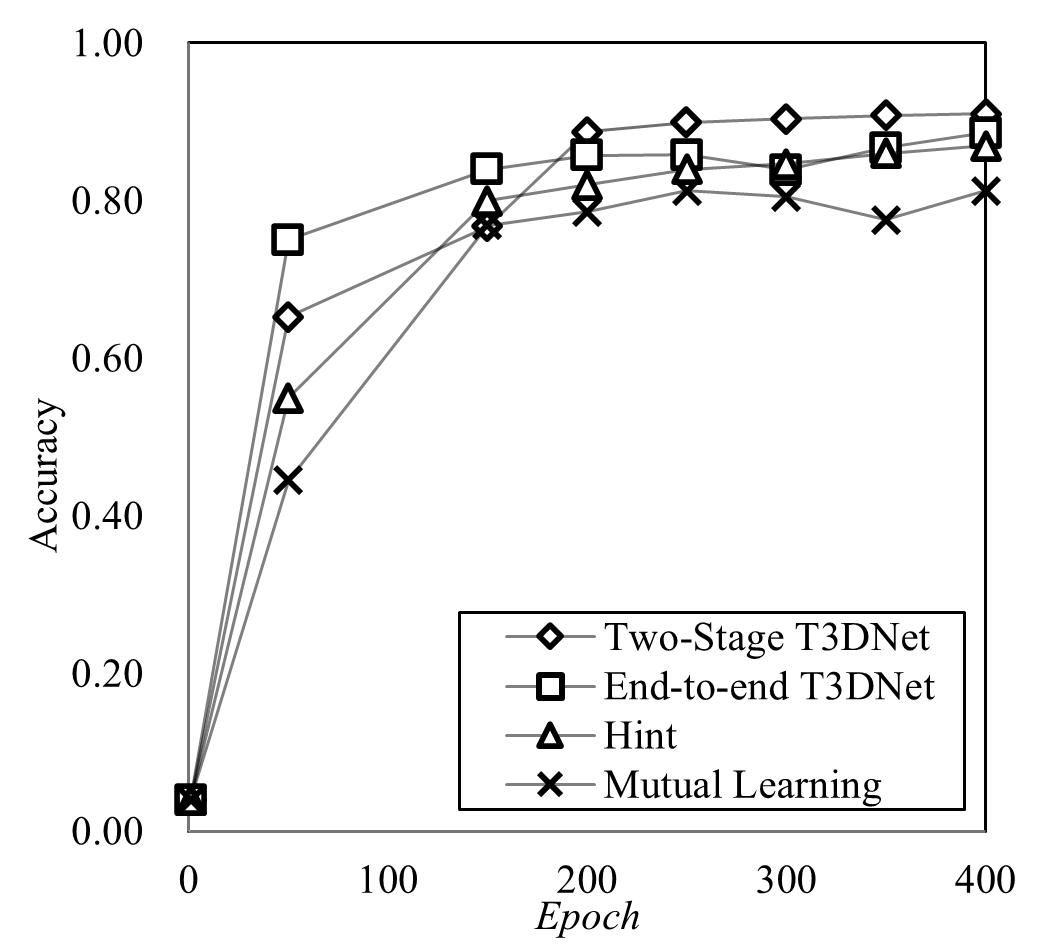}
\caption{Validation set accuracy curve of different distillation methods attempts in the training procedure. The horizontal and vertical axes are epochs and the validation accuracy, respectively.}
\label{fig:training_curve} 
\end{figure}

Table~\ref{table:Distill} summarizes the results of the ablation studies. We use a 1/64 tiny model as the baseline. We can see that the feature-level distillation loss (hint) has the test accuracy of 86.99$\%$. It hurts the performance by 1.37$\%$ instead of improving it. The mutual learning method(81.25$\%$) also significantly descends the performance by 7.12$\%$. It shows that these distillation techniques can not help the learning of the tiny PointNet++ model. The end-to-end strategy only has 88.34$\%$ test accuracy, which even hurts 0.02$\%$ accuracy of the original model instead of improving it. There are chances that the failure of the end-to-end training method is due to the improper setting of the hyperparameters. When we analyze the underlying reason, we find that the composition of the total training loss is too complex. The training is bound by two different kinds of loss at the same time, which makes the network difficult to converge. JMC~\cite{cui2021joint} is a method that combines KD with another compression technique: pruning. It explains that the network after pruning hurts performance little by little, thus KD can recover performance right after pruning during the training stage. Different from JMC~\cite{cui2021joint}, our method directly defines a tiny model from scratch and uses the T3DNet to tackle the under-fitting problem. The framework of end-to-end learning may not be suitable in this situation. Among all these methods, only the two-stage T3DNet improves the test accuracy to 91.01$\%$. Through exploratory studies, we can prove that the two-stage T3DNet is our best design.

\textcolor{black}{Fig.~\ref{fig:training_curve} depicts the validation set accuracy curve for various distillation methods during the training process. Notably, the two-stage T3DNet method exhibits a gradual ascent during the initial 200 epochs (stage 1). This gradual progression is attributed to the shared weights in the largest augmented models during tiny network augmentation, where different augmented models are trained at each epoch. Consequently, the convergence is tempered, leading to a more gradual increase in accuracy. Nevertheless, in the subsequent 200 epochs (stage 2), the tiny network distillation achieves higher accuracy compared to other distillation attempts. This observation aligns with our findings, suggesting that the model's responsiveness to knowledge distillation is enhanced after network augmentation.}

\textcolor{black}{The training curves for the other three distillation attempts are also presented in Fig.~\ref{fig:training_curve}. The end-to-end strategy converges rapidly initially but becomes plateaued in accuracy as training progresses. Both hint and mutual learning strategies also fall short of achieving satisfactory accuracy levels during the training procedure.}

\begin{table}[htbp!]
\centering
\caption{Ablation study on KD with different $T$}
        \begin{tabular}{c|c|c}
            \toprule
            Model(\#Params=0.03M, FLOPs=589M) & Test Acc &$\Delta$ Acc\\
            \midrule
            Tiny model (Baseline)&88.37 &-   \\
            \midrule
            Tiny model + KD ($T=1$)& 89.39 & +1.09\\
            Tiny model + KD ($T=2$)& 88.99 & +0.63\\
            Tiny model + KD ($T=5$)& 87.74 & -0.62\\
            Tiny model + KD ($T=10$)& 87.66 & -0.70\\
            Tiny model + KD ($T=20$)& 88.47 & +0.10\\
            \bottomrule
        \end{tabular}
    % \end{minipage}}
\label{table:Different_T}
\end{table}

Table~\ref{table:Different_T} summarizes the experiments of different $T$ used in the knowledge distillation. When we use $T=1$ in the KD, we can get the best performance of 89.39$\%$, improving 1.09$\%$ compared with the baseline(88.37$\%$). $T$=2 gives the test accuracy of 88.99$\%$, the increment(0.63$\%$) is slightly smaller than that of $T=1$. When we turn to larger $T=5, 10, 20$, the test accuracy is 87.74$\%$, 87.66$\%$, and 88.47$\%$ respectively. The experiments show that a larger $T$ does not give better performance. Generally speaking, a higher $T$ means softer distilled knowledge, which leads to better performance in most cases~\cite{hinton2015distilling}. However, the result shows that in our studies, $T=1$ gives the best performance. A large $T$ makes the distillation process unstable and even hurts the performance in some cases. Therefore, it is recommended to use a value of $T=1$ for the experiments.

\section{Conclusion}
\label{sec:conclusion}
% Driving by the flourishing of autonomous vehicles and 3D sensing, there is a stringent need to push a point cloud learning model to be deployed on the edged devices. Thus, it is crucial to compress powerful models to fulfill the computational and memory constraints of the microcontroller or edged GPU without a significant accuracy drop. Although many existing works design efficient inference components to speed up the learning process, to the best of our knowledge, no one has applied model compression techniques directly to the 3D point cloud models.

In this paper, we conduct the model compression from another perspective: pre-define a tiny model out of the original one and tackle its under-fitting problem to improve its accuracy. We propose a framework called tiny network with augmentation and distillation (T3DNet). The method adopts a two-stage strategy. We find that the tiny model after network augmentation is better for distillation. We realize the network augmentation by embedding the target tiny model into the original model to get auxiliary supervision by randomly expanding the layer size. And a pre-trained original model works as the teacher for the distillation stage. Compared with augmentation or distillation only, our two-stage T3DNet can achieve the best accuracy improvement. The method is verified on the famous PointNet++ model using ModelNet40, ScanObjectNN, and ShapeNet. Finally, we build a 58$\times$ smaller tiny network with only a 1.45$\%$ insignificant accuracy drop, and 2.64$\%$ accuracy improved compared with the pure tiny model without T3DNet on ModelNet40. We also conduct exploratory experiments to attempt different kinds of distillation methods and combinations. 

We focused our studies on the 3D object classification task and 3D part segmentation task in this paper. However, T3DNet is designed to be a generic method, we hope to apply it to other datasets such as ScanNet~\cite{dai2017scannet} in the future. It can also be applied to other tasks like 3D object detection and 3D semantic segmentation. Besides, despite the failure of our exploration of different distillation and combination strategies, we hope to continue searching for ways to improve the T3DNet framework in the future.

\bibliographystyle{IEEEtran}
\bibliography{bio.bib}

\end{document}